# Language Distribution Prediction based on Batch Markov Monte Carlo Simulation with Migration

**XingYu Fu**    **ZiYi Yang**    **XiuWen Duan**

Sun Yat_sen University

{fuxy28, yangzy7, duanxw3}@mail2.sysu.edu.cn

## Abstract

Language spreading is a complex mechanism that involves issues like culture, economics, migration, population etc. In this paper, we propose a set of methods to model the dynamics of the spreading system. To model the randomness of language spreading, we propose the *Batch Markov Monte Carlo Simulation with Migration(BMMCSM)* algorithm, in which each agent is treated as a *language stack*. The agent learns languages and migrates based on the proposed *Batch Markov Property* according to the transition matrix $T$ and migration matrix $M$. Since population plays a crucial role in language spreading, we also introduce the *Mortality and Fertility Mechanism,* which controls the birth and death of the simulated agents, into the BMMCSM algorithm. The simulation results of BMMCSM show that the numerical and geographic distribution of languages varies across the time. The change of distribution fits the world's cultural and economic development trend. Next, when we construct Matrix $T$, there are some entries of $T$ can be directly calculated from historical statistics while some entries of $T$ is unknown. Thus, the key to the success of the BMMCSM lies in the accurate estimation of transition matrix $T$ by estimating the unknown entries of $T$ under the supervision of the known entries. To achieve this, we first construct a $20 \times 20 \times 5$ factor tensor $\vec{X}$ to characterize each entry of $T$. Then we train a Random Forest Regressor on the known entries of $T$ and use the trained regressor to predict the unknown entries. The reason why we choose Random Forest (RF) is that, compared to Single Decision Tree, it conquers the problem of over-fitting and the Shapiro test also suggests that the residual of RF subjects to the Normal distribution.

# 1 Mechanisms for Learning Languages

Supposing there are $N$ languages $L = \{l_1, l_2, \ldots, l_N\}$ to be modelled, we treat $L$ as the states set in the classical Stochastic Markov theory where each agent learns new language $l_j$ based on the batch of languages he(her) has already mastered and the transition matrix $T = (t_{ij})_{N \times N}$ of $L$, in which:

$$t_{ij} = P\{agent\ will\ learn\ l_j\ |\ agent\ has\ mastered\ l_i\}.$$

The construction of transition matrix $T$ depends on a thorough research on a selection of factors which determine the learning trend for different languages speakers. Some Regression techniques are applied to make an accurate prediction of each $t_{ij}$. We will cover the details of the construction in section 4.

One thing to note is that, the Markov property in traditional Stochastic Markov theory [1]:

$$P\{X_{t+1} = i | X_t, X_{t-1}, \ldots, X_0\} = P\{X_{t+1} = i | X_t\}$$

no longer holds in our case, and therefore we propose the *Batch Markov Property(BMP)* in which the next language each agent is going to learn is dependent on the whole batch of languages he(her) has already mastered. Figure1 illustrates the intuition of BMP:

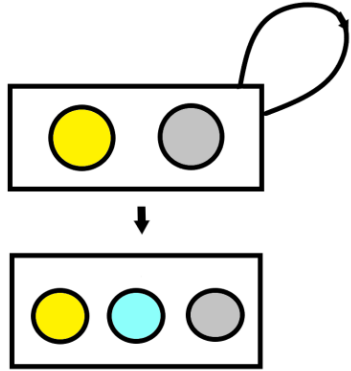

*Figure1: a batch of learned languages behaves as an integrated unit*

To make the concept of BMP clear, imagine $L = \{l_1, l_2, l_3\}$, e.g. $\{Korean, English, German\}$, and there is a specific agent who has already mastered languages $l_1$ and $l_3$. We now calculate the probability of the event that the agent masters $l_2$ in the next round by assuming that the agent is currently uniformly distributed among the batch $\{l_1, l_3\}$, by total probability rule [2], that is:

$$P\{agent\ will\ learn\ l_2\ |\ agent\ has\ mastered\ l_1\ and\ l_3\} = (t_{12} + t_{32})\ /\ 2.$$

Another mechanism of interest in our work is the *Language Stack Mechanism(LSM)*, where we imagine each learning agent as an ever-updating stack which is pushed into a language at each round while the layer is deleted from the top if it turns out to be the same language as another already inserted layer. Figure2 shows an example of LSM:

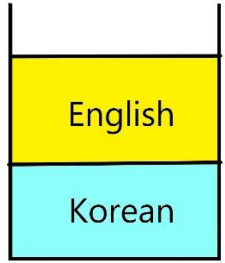

*Figure2: an agent can be deemed as an ever-updating stack*

The $n^{th}$ layer of the stack represents the $n^{th}$ language of the agent from the bottom up (first layer is native language) and the deleting mechanism guarantees that the agent can stop learning new languages, which is quite common for most of us, and there are no repeated languages in the language stack for each agent

# 2   Batch Markov Monte Carlo Simulation

To study the distribution of various language speaker over time, we propose the *Batch Markov Monte Carlo Simulation(BMMCS)* algorithm to model the collective learning behaviors of a sample of agents as time develops.

We run the algorithm multiple times (100 times, in our case) with rather large initial sample number (1 million, in our case). By the Law of Large Numbers [2], we form a stable and accurate statistical prediction of the future languages distribution.

In our Monte Carlo Simulation, we first sample an initial group of language learning agents whose languages subject to the initial distribution of various languages. [18]

At each year, some of the sampled agents will learn new languages based on the BMP mechanism that we have just discussed in section1 and some of the sampled agents will not learn new languages, which is rather common since language learning process is painful for most of us. By simulating like this, we can see the dynamics of how languages spread over a group of people whose first and second languages are quite diverse.

One phenomenon we need to pay special attention to is that some languages that are not that international are actually spoken by a large number of population, e.g. Chinese (Rank 1st in native languages) and Hindi (Rank 3rd in native languages) [3]. The reason for this phenomenon is obvious since it turns out China and India are the countries of most population on Earth [4] and therefore such languages are spoken most. Figure3 visualize the distribution of world population by country in 2017 [4]

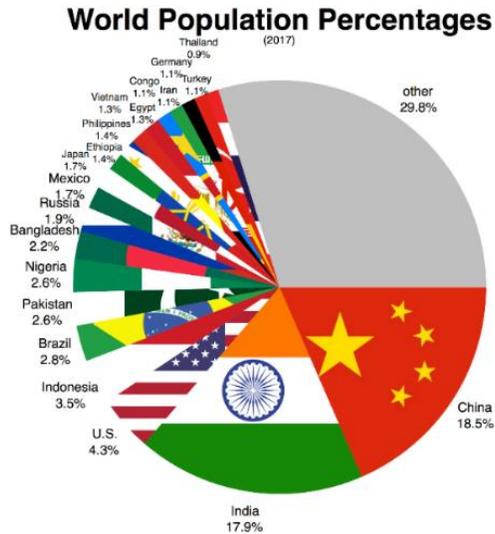

*Figure3: World Population Percentage by country*

From above, we can see that the factors closely related to population distribution and population distribution itself can influence the languages distribution profoundly. Hence, we adopt two methods to take this into consideration. First, the initial sampling has already contained the population information. Second, we implement the *Mortality and Fertility Mechanism (MFM)* in the BMMCS algorithm to simulate the dynamics of population distribution over time. The parameters involved in MFM is described below:

We download the database of the prediction of the population pattern of each country in the coming 50 years which is modeled and predicted by the Department of Economic and Social Affairs of the UN [19]. We take 5 years as a term (10 terms for 50 years) and summarize countries' population of each term to get each term's language zone population distribution. When running BMMCS, we update the population through the time to get a more practical stimulation. Figure4 shows the pipeline of BMMCS algorithm:

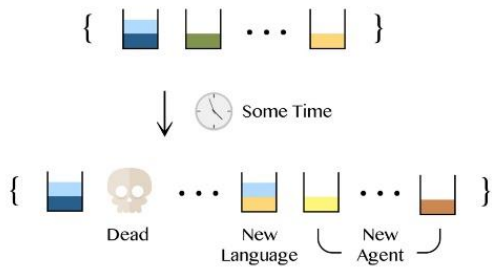

*Figure4: the pipeline of BMMCS*

To summarize the discussion above, we write the algorithm here:

**Algorithm1 (BMMCS)**:

# Establish the initial society

- Sample an initial group of language stacks (learning agents) according to the initial distribution of various languages at the starting time.

**for** the length of time **do**

# BMP Learning

- Let each stack learns language based on the transition matrix $T$ and the BMP mechanism.

# MFM

- Delete a proportion of stacks based on the death rate $\alpha_0$ of different zones in that time.
- Add new stacks into the system whose native languages (first layer of stack) subject to the birth rate $\alpha_1$ and population of different language regions in that time.

**end for**

# 3 Batch Markov Monte Carlo Simulation with Migration

In section 2, we proposed the so-called Batch Markov Monte Carlo Simulation algorithm where each agent (language stack) is learning new languages as time develops and some of the agent may pass away while some new agent may be born by the mortality and fertility mechanism we proposed in BMMCS. To summarize, BMMCS takes the spread of languages and the population changes into consideration.

However, BMMCS doesn't include the human migration, which is intuitively crucial for the geographic distribution of languages, and therefore we add the migration mechanism in this section. We name the final simulation strategy which consider human migration as *Batch Markov and Monte Carlo Simulation with Migration (BMMCSM).*

In BMMCSM, we first construct the migration preference matrix $M = (m_{ij})_{N \times N}$, where:

$m_{ij} = P\{\text{agent will migrate to language zone } j \mid \text{agent stay in language zone } i\}$.

While, note that the migration pattern for each agent relies on not only the agent's current living location but also has a strong connection to the mother land (i.e. the language zone of the native language) of the agent. Thus, for an agent whose mother land is in language zone $k$ and currently living in language zone $i$, we calculate the distribution of his(her) next living language zone as following:

$$(m_i + m_k)/2.$$

, where $m_i$ denotes the $i_{th}$ row of $M$. The intuition for the above equation is that the agent uniformly distributes between his(her) birth place and currently living place by the total probability rule [2].

To summarize the above discussion, we write the algorithm here:

**Algorithm2 (BMMCSM)**:

# Establish the initial society

- Sample an initial group of language stacks (learning agents) according to the distribution of various languages at the starting time. We assume the current location for each agent is his(her) mother land.

**for** the length of time **do**

# BMP Learning

- Let each stack learn language based on the transition matrix $T$ and the BMP mechanism.

# Migration (Difference between BMMCS and BMMCSM)

- Each agent changes their current living location subjecting to $(m_{mother} + m_{current})/2.$

# MFM

- Delete a proportion of stacks based on the death rate $\alpha_0$ of the time.
- Add new stacks into the system whose native languages (first layer of stack) subject to the birth rate $\alpha_1$ and population of different language regions of the time.
- 

**end for**

After the simulation, we end up with a group of agents spread over each language zone and we know exactly what languages are in the language stack for each agent. Hence, we can know the languages distribution for each language zone after years' migration, languages spread, and population change.

# 4 Estimation of transition matrix $T$

## 4.1 Sketch of Section 4

So far, we have shown how the BMMCS and BMMCSM algorithms work under the assumption that the transition matrix $T = (t_{ij})_{N \times N}$ has been given. In this section, we will show the pipeline of building matrix $T$ specifically.

To recap, we write the definition of $t_{ij}$ here again:

$$t_{ij} = P\{agent\ will\ learn\ l_j\ |\ agent\ has\ mastered\ l_i\}$$

For some $t_{ij} \in T$, we can tell the value of it directly from historical statistical data, while for some other $t_{ij} \in T$, this value can't be known directly. Therefore, we result in a sparse matrix $T$ like this:

$$\begin{pmatrix} 0.9 & 0.1 \\ 0.2 & \end{pmatrix}$$

What remains to be solved is estimating the unknown $t_{ij}$ from the known ones, which falls into the classical supervised learning paradigm.

To implement regression, for each $t_{ij} \in T$, we look for factors that determine the willingness of language $i$ speakers to learn language $j$. Say we have $n$ factors $(x_{ij}^{(1)}, x_{ij}^{(2)}, \ldots, x_{ij}^{(n)})$ for $t_{ij}$, then we want to find a decision function $f: R^n \to R$ such that:

$$t_{ij} = f(x_{ij}^{(1)}, x_{ij}^{(2)}, \ldots, x_{ij}^{(n)}) + \varepsilon_{ij}\ ,\ \forall (i,j) \in L \times L$$

where $\varepsilon_{ij}$ is a Normal(Gaussian) random noise:

$$\varepsilon_{ij} \sim N(0, \sigma^2)$$

In our work, we compare two models to approximate decision function $f$, i.e. *Classification And Regression Tree(CART)* [5] and *Random Forest(RF)* [6], which turns out that CART results in serious over-fitting problem while RF can elegantly overcome it by its Random Sampling, Random Split and Ensemble Mechanism [6]. The comparison will be discussed in detail in section 1.4.4. Figure5 visualize the pipeline of Random Forest learning algorithm [7]:

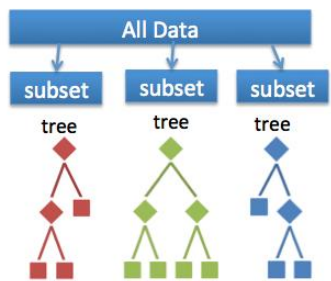

*Figure5: Random Forest*

We train our model on the set of known $t_{ij}$(training dataset), and then predict the unknown $t_{ij}$ by the already well-trained model. The construction of matrix $T$ is completed here.

The key to the success of our estimation lies in the choice of factors which will be explained in detail in next sub-section.

## 4.2   Construction of Factors

There are many aspects of factors influence the willingness of language $i$ speakers to learn language $j$. To well estimate the unknown $t_{ij}$, we now introduce the factors that we choose in this work. It is also natural to add more factors into the framework which can be studied as future research.

**Set Up: Classification of Language Zone and Data Pre-Preparation**

We studied 26 most used languages and clustered them into 20 language zones. Here, we give each language zone an index, that is:

$$L = \{l_0, l_1, \ldots, l_{19}\}$$

All data analyzed in our model is collected in terms of language zones.

Among the 26 languages, Mandarin Chinese, Wu Chinese and Yue Chinese are putted together as Language zone *Chinese* while Telugu, Tamil, Marathi and Punjabi are considered as sub branch of *Hindi/Urdu* Language family. The reason behind this clustering is that these languages are most spread and spoken in a relatively close geographical area and the developments of these languages are mostly influenced by the characteristics of the same countries and districts.

To construct the factors and transition matrix $T$, we collect data of different countries concerned their official language, economic status and culture influence. Then, we match the countries with their corresponding language zones and further processed the data to calculate our factors and then use the factors to estimate the transition matrix $T$.

As for the countries we chosen, we consider the top 30 most populous countries and the top 30 countries with strongest culture soft power, since they are the most influential countries in the language zones. The figure below shows the countries that we study for each language zone:

| No. | language | Main Country | No. | language | Main Country | No. | language | Main Country |
|---|---|---|---|---|---|---|---|---|
| 0 | Mandarin Chinese | China | 3 | Spanish | Mexico | 10 | Hausa | Africa |
| 1 | English | US | | | Colombia | 11 | Japanese | Japan |
| | | Nigeria | | | Spain | 12 | German | Germany |
| | | Philippines | 4 | Arabic | Egypt | | | Switzerland |
| | | UK | | | Ethiopia | 13 | Persian | Iran |
| | | South Africa | 5 | Malay | Malaysia | 14 | Swahili | Tanzania |
| | | Myanmar | 6 | Russian | Russia | 15 | Javaness | Indonesia |
| | | Kenya | 7 | Bengali | Bangladesh | 16 | Korean | South Korea |
| | | Canada | 8 | Portuguese | Brazil | 17 | Turkish | Turkey |
| | | Australia | | | Portugal | 18 | Vietnamese | Vietnam |
| 2 | Hindustani | India | 9 | French | Congo | 19 | Italian | Italy |
| | | Pakistan | | | France | | | |

*Figure6: 35 countries are selected and assigned to 20 language zones.*

**Factor1: Language similarity**

Firstly, based on the Ethnology theory, *Language Similarity* is an important motivation for people to acquire a new language, especially when the new language that the people try to learn shares similar analogical syntax or grammar system with the language they have already mastered [11]. For example, it is a common phenomenon in Europe that a person who takes German as his native language prefer to learn Dutch than other people due to the similarity of these two languages. We Thus, we introduce the factor $Simi\ (i,j)$ to quantify any two languages' similarity in the field of Ethnology Language family.

According to the data in appendix, most languages in our 20 language zones belong to 8 language families, excluding Japanese and Korean. (Japanese and Korean are two relatively independent languages. [12]) Then, in our paper, we use the method of Hot-Encoding by assigning each language zone a vector with 8 dimensions representing 8 language families. For each language zone $i$ 's language family vector, we assign 1 to the language families it belongs to and 0 to other dimensions. While for Japanese and Korean, we assign their vectors with $\vec{0} = (0,0,0,0,0,0,0,0)$.

Given any two languages $i,j \in L$, their similarity can be calculated by inner product, that is:

$$Simi\ (i,j) = \vec{i}\ \vec{j}$$

where $\vec{i}$ and $\vec{j}$ represent the language family vectors for language $i$ and $j$ respectively.

If $Simi\ (i,j) = 1$, we say these two languages share similarity, otherwise we say the two have no similarity.

## Factor2: Foreign Direct Investment Net Outflows(FDINO)

The openness of countries can influence the popularities of their languages. We use the amount of one zone's *Foreign Direct Investment Net Outflows(FDINO)* to quantify the openness. From the FDINO, we can see the government's willingness to communication with other regions, so we can call it the extroversion of one language zone. The data we use is from the World Bank [13] [14], while the data were distinguished by countries. Thus, we use the above formulas to get the data we want (i.e. distinguished by language zones), that is:

$$extro(i) = \sum_{a \in i} export(a) \cdot weight(a), i \in L$$

$$weight(a) = \frac{GDP \ of \ country \ a}{GDP \ of \ language \ zone \ i}$$

$$(GDP \ of \ language \ zone \ i = \sum_{k \in i} GDP \ of \ country \ k)$$

Further, for any $i, j \in L$, we assume that $extro(i,j) = extro(j)$, which makes sense since the willingness of learning language $j$ is mostly dependent on the situation of $j$, rather than $i$.

## Factor3: Economic Interaction between Language Zones

As global communication increases, there emerges international business and global tourism, which can impose influence on the spread of languages. We need to consider the *interaction* between language zones as a factor, to illustrate, if one zone intends to do business with another zone, then the people from the former zone have more motivation to learn the latter zone's language, which is necessary for the buyer and the seller to communicate.

Therefore, we use the *Export Percentage* from one zone to another zone measure the importance the former imposes on the latter, defined as $export(i,j), i,j \in L$. If this factor is high, means that region $i$ has a large amount of export towards region $j$, then region $j$ seems more connected to region $i$, and therefore there is a tendency for people in region $i$ to learn the language in region $j$.

The data we use to calculate $export(i,j)$ is from a famous official business website in China [15], and this dataset were based on Department of Statistics of each country. However, we can only obtain data between countries, so we need to convert it into data between language zones.

Therefore, we calculate as follow:

$$export(i,j) = \sum_{a \in i} \sum_{b \in j} export(a,b) \cdot weight(a)$$

$$weight(a) = \frac{GDP \ of \ country \ a}{GDP \ of \ language \ zone \ i}$$

Figure7 illustrates the intuition of this calculation:

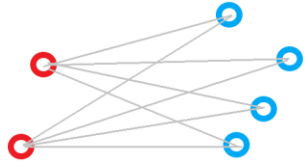

*Figure7: the circles with the same color represent countries in the same language zone and the Economic Interaction between two language zones are calculated by summing up the exports between all possible pairs*

**Factor4: Culture Soft Power**

Culture soft power is also a key factor we considered in our model. Language of high culture export countries enjoys higher language exposure rate. For these languages, foreigners can get access to them easily through culture products like music, TV drama, movies and so on. Even some pop culture may cause a hit to learn certain language. For example, many teenagers all over the world have been attracted by Korean pop culture or Japanese anime to learn these countries' languages. A rank with score officially provided by USC Center on Public Diplomacy [16] shows us each country's soft power. Summarize the scores so that we get the assessment of language zones' soft power ($softpower(i), i \in L$). Here we assume that countries inside each language zone contribute to the spread of its language independently and additively.

Since each zone's culture exports to countries all over the world, regardless of the speakers' original zones, we therefore assume that $soft(i,j)$ is independent of $i$, that is:

$$soft(i,j) = softpower(j), for\ \forall i,j \in L$$

**Factor 5: Migration Preference**

The immigration and emigration patterns are good tools for us to predict the future distribution of language. In our predictive model, we introduce *Migration Preference* which represents the probability that people immigrate into another language zone from one language zone. Migration Preference is derived from the 2017 the UN workbook of Migrant Stock by Origin and Destination [17]. We summarize corresponding countries' data and know how many people migrate from language zone $i$ and finally settle in language zone $j$, which we denote as $e_{ij}$.

Then, we calculate each entry $m_{ij}$ of *Migration Matrix M*, which represents the probability people immigrate into language zone $j$ from language zone $i$, by the following formula:

$$m_{ij} = \frac{e_{ij}}{\sum_{k \in L} e_{ik}}, i,j \in L$$

We not only consider $m_{ij}$ as factor for predicting $t_{ij}$, but also use it to simulate the population migration in BMMCSM.

## 4.3 Estimation of $T$

For each pair $(i,j) \in L \times L$, we construct the corresponding factor vector as we discussed in section 4.2 :

$$\vec{X}_{ij} = (x_{ij}^{(1)}, x_{ij}^{(2)}, \ldots, x_{ij}^{(5)})$$

and therefore, we end up with a $20 \times 20 \times 5$ factor tensor $\vec{X}$. Figure8 visualizes the structure of $\vec{X}$:

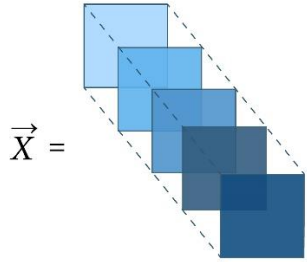

Figure8: Factor Tensor $\vec{X}$

To calculate all the entries in $T$, we first calculate the $t_{ij} \in T$ that can be directly known from historical statistics, which end up with a sparse estimation of the transition matrix $T$, i.e. many of the entries in $T$ are empty. The work we left now is to make an inference for the empty entries in $T$, which can't be directly estimated, based on the known $(\vec{X}_{ij}, t_{ij})$, i.e. the supervised learning paradigm. Formally, we define the training dataset and predicting dataset as follow:

$$\vec{X}_{train} = \{ (\vec{X}_{ij}, t_{ij}) \mid t_{ij} \text{ is directly estimated from historical data}\}$$

$$\vec{X}_{predict} = \{ \vec{X}_{ij} \mid t_{ij} \text{ is empty }\}.$$

We train our models (CART and RF) on the training dataset $\vec{X}_{train}$ to approximate the decision function $f$, which end up with two regressors: $f_{CART}$ and $f_{RF}$. Then we apply the two regressors to $\vec{X}_{predict}$ and completes the inference of transition matrix $T$. Figure9 shows our estimated $T$ predicted by Random Forest (with 100 trees and maximal depth of 4 to avoid over-fitting):

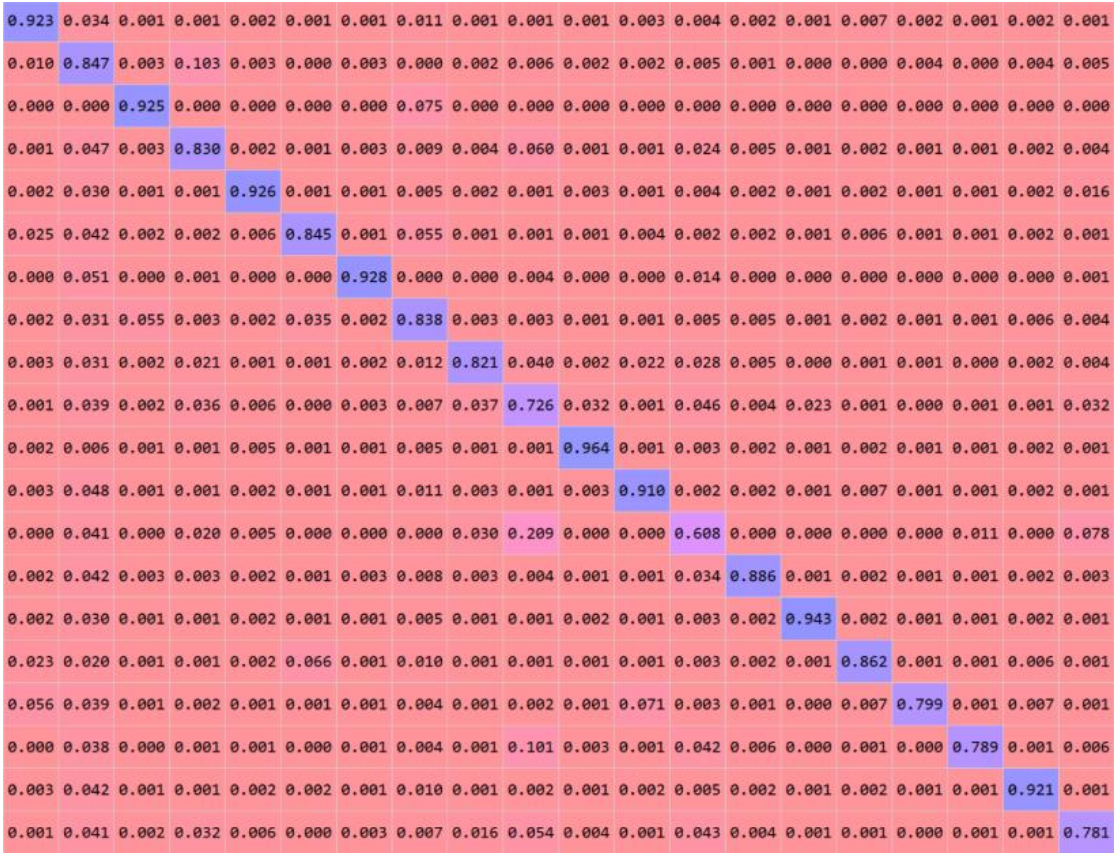

*Figure9: T predicted by Random Forest*

In our work, we use scikit-learn [10], a popular open sourced implementation of many machine learning algorithms in Python programming language community, to apply RF and CART efficiently.

## 4.4 Result Analysis

We calculate the residual for both $f_{CART}$ and $f_{RF}$ on the training dataset respectively by:

$$\overrightarrow{Residual} = \{ \varepsilon_{ij} = (t_{ij} - f(\vec{X}_{ij})) | (\vec{X}_{ij}, \varepsilon_{ij}) \in \vec{X}_{train}\}.$$

Now we analyze the following two questions: 1) Can we estimate the generalization abilities for the two models? 2) Does the residual subjects to the normal distribution? The former question handles the tradeoff between bias and variance [8], which is a tricky problem in the Machine Learning research community. The latter one is an examination of our statistical assumption that:

$$t_{ij} = f(x_{ij}^{(1)}, x_{ij}^{(2)}, \ldots, x_{ij}^{(n)}) + \varepsilon_{ij} \ , \forall (i,j) \in L \times L$$

where:
$$\varepsilon_{ij} \sim N(0, \sigma^2)$$

We draw the histograms of $\overrightarrow{Residual}_{RF}$ and $\overrightarrow{Residual}_{CART}$ below:

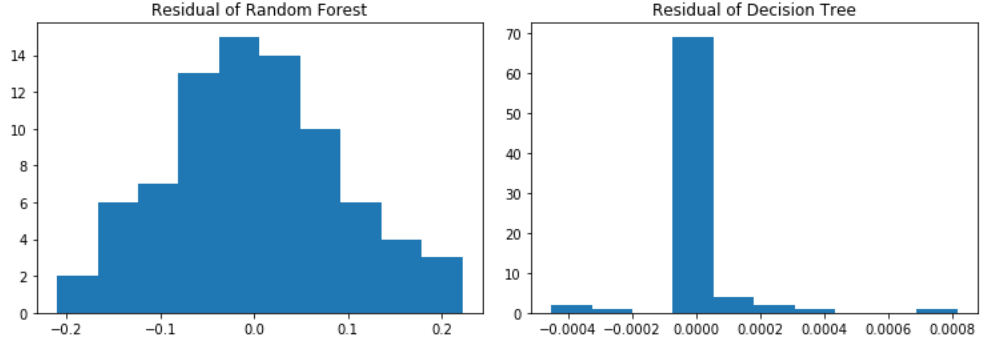

*Figure 10: Histograms of residuals for RF and CART*

We can directly see from the above figure that the residuals for RF roughly subject to the normal distribution while the residuals for CART are almost 0, which means that CART over-fits the dataset and loses the ability of generalization. The RF conquers the problem of over-fitting by its clever designs of Random mechanism and Ensemble mechanism.

To further exam the normality of $\overrightarrow{Residual}_{RF}$, we apply the Shapiro-Wilk test [9] to the residual vector of RF and the $p-value$ turns out to be 0.23942, which is above the significance level 5%. Thus, we accept hypothesis $H_0$ that $\overrightarrow{Residual}_{RF}$ is normally distributed, which exams the validity of our model.

# 5 Simulation Results

Section 1 to Section 4 describe specifically our methodology for modelling the distribution of various languages over time dimension and over geography dimension. To summarize, we use Random Forest Machine Learning technique to estimate the parameters (transition matrix $T$) involved in our system, and then run the proposed Batch Markov Monte Carlo Simulation with Migration(BMMCSM) algorithm to simulate the dynamics of language as time develops.

Our simulation result shows that the numerical and geographical distribution of different languages will change in an insightful way in the future, which will be demonstrated in the following sub-sections.

## 5.1 Numerical Distribution

In the next fifty years, the distribution of different language speakers will change notably in several aspects including the total speakers number, first language speakers number and second language speakers number based on our numerical simulation. Figure11 shows the future change of distribution of total language speakers (including first, second, and third speakers etc.).

*(For Figure 11,13, and 14. From left to right are: Chinese, English, Hindi, Spanish, Arabic, Malay, Russian, Bengali, Portuguese, French, Hausa, Japanese, German, Persian, Swahili, Javanese, Korean, Turkish, Vietnamese, Italian)*

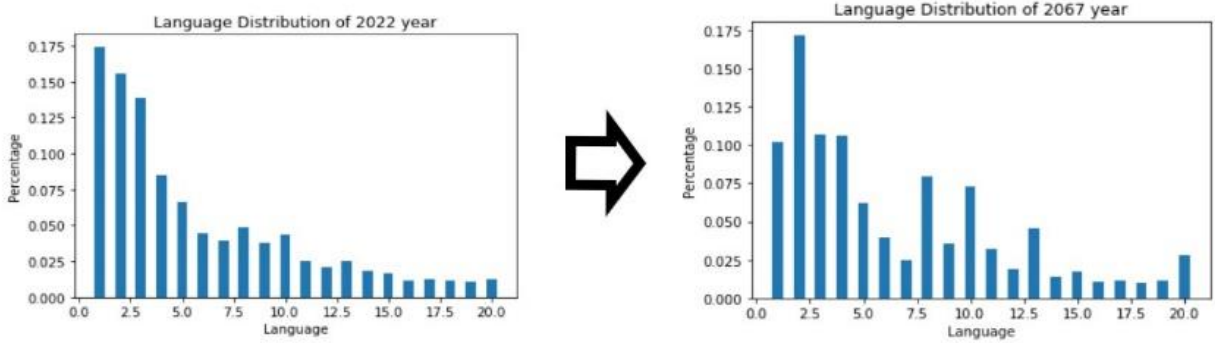

*Figure11: Change of total language speakers in the next fifty years*

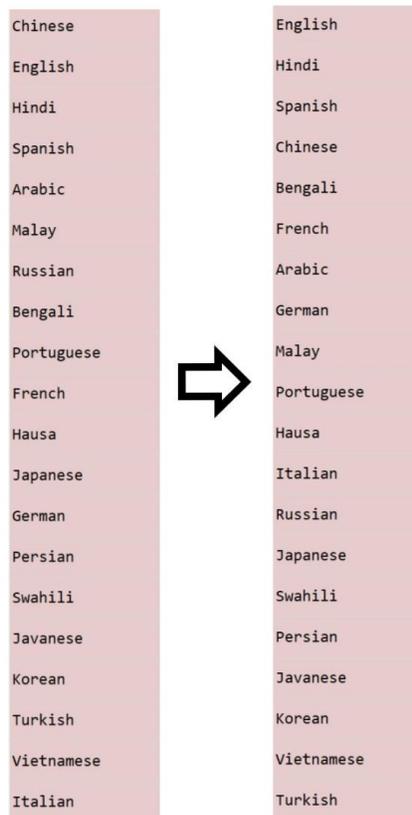

*Figure12: Change of Language Ranking*

As for total speaker numbers, Figure 11 and Figure 12 clearly shows the Language Ranking change in the next 50 years. Compared with initial top 10 most used language, the rank changes while the components stay relative stable. English becomes the most common used language in the word which may reveal the further progress of globalization with aspects of electronic communication and social media. The rank change of Chinese and Hindi are mainly due to the population pattern. Moreover, German takes place of Russian to be one of the top 10. It may be explained that Germany may embrace high- speed development period after the Brexit.

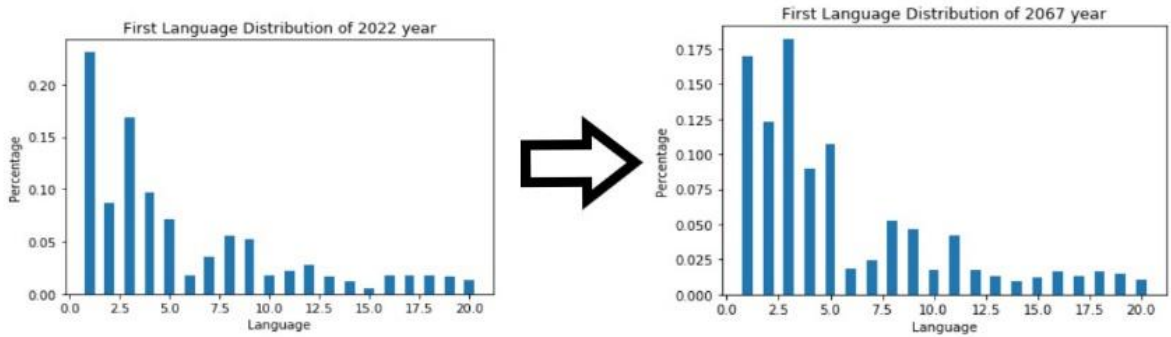

*Figure13: Change of first language speakers in the next fifty years*

As for first language speaker, Hindi ranks the first while Chinese drops to the second place. It meets the future population trend. According to the world population pattern, China will reach its population peak in next 10 years and then decline. Meanwhile, India will exceed China with a continuously population growth.

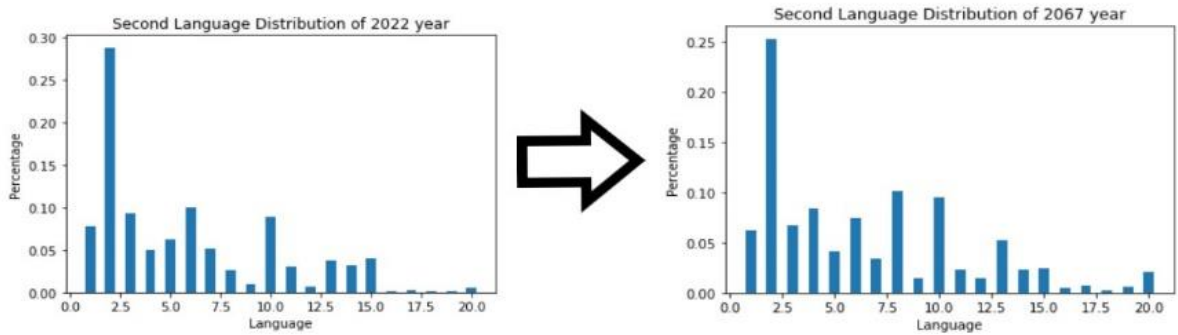

*Figure14: Change of second language speakers in the next fifty years*

As for second language, English remains to be the most popular choice since English is regarded as an official language in transnational communication. The rise of other language second speakers also indicates that much more people would like to acquire a second language and their choices are diversified.

## 5.2 Geographic Distribution

With the help of Batch Markov Monte Carlo Simulation with Migration algorithm, we can further study the geographic distribution of languages when take human migration into consideration.

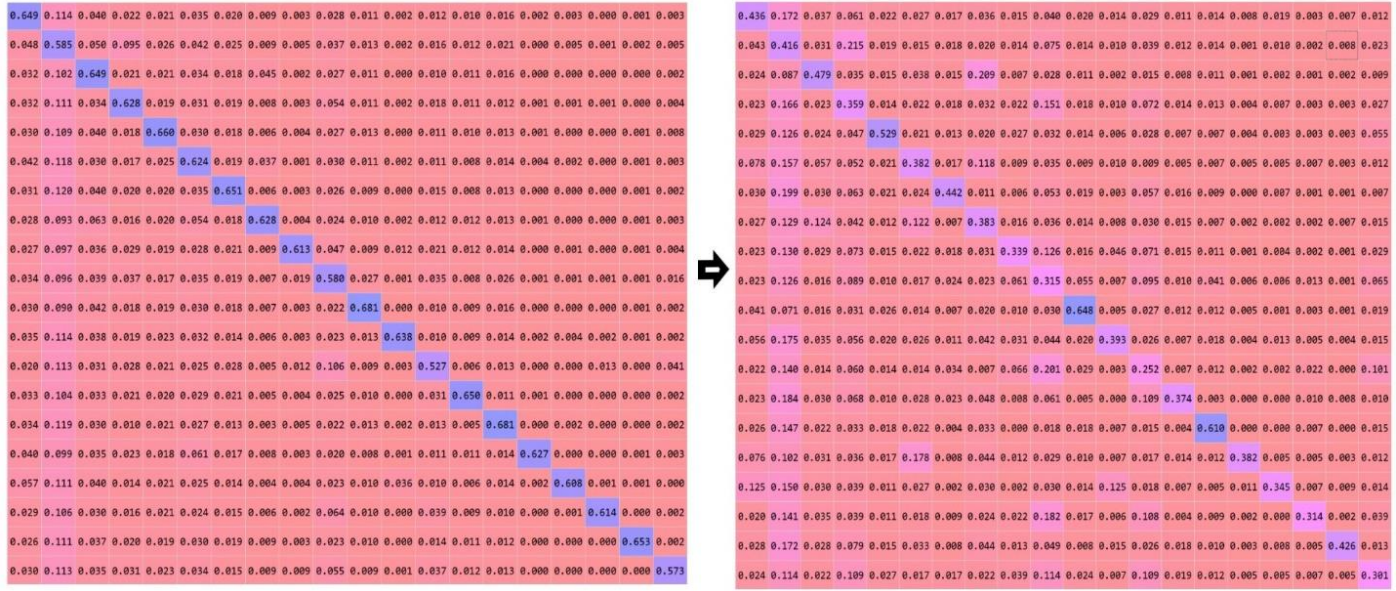

*Figure15: Change of geographic language distribution*

The two figures above demonstrate the geographic distribution of languages before or after the migration. The value in the $i_{th}$ row and $j_{th}$ column of the matrix means the percentage of $j$ language speakers of total speakers living in language zone $i$. The color intensity of the $i_{th}$ row and $j_{th}$ column of the matrix shows the percent of how many people in this area (language zone $i$) can speak language $j$ . So, as we can see from the left part of figure15, for any $j$, the speaker of $j$ centralizes on the diagonal line of the matrix, i.e. language zone $j$. But in the right part of figure15, we can see the distribution becomes sparse, which implies the migration of people advances the cultural collision and spread their native languages. To be specific, as there is a lot of people flooded into the English language zone, this district reveals high language diversity. Languages in Europe tend to assimilate each other and enlarge the speakers of each language. Some languages spread to different countries of the world while they might be spoken only within a small region previously.

Appendix1: Population trend in next 50 years

Population trend of language zones in next 50 years

| Language zone | No | 2017 | 2022 | 2027 | 2032 | 2037 | 2042 | 2047 | 2052 | 2057 | 2062 | 2068 |
|---|---|---|---|---|---|---|---|---|---|---|---|---|
| Mandarin Chinese | 0 | 1388232693 | 1409374980 | 1416294130 | 1413475690 | 1403622846 | 1387474164 | 1364798264 | 1335637565 | 1300266694 | 1260738881 | 1213021592 |
| English | 1 | 916043460 | 977215346 | 1039241875 | 1101685176 | 1164231765 | 1226677990 | 1288740374 | 1350221817 | 1410886890 | 1470582285 | 1540220888 |
| Hindustani | 2 | 1539257082 | 1634743618 | 1723319989 | 1803658275 | 1874717041 | 1936241176 | 1988377093 | 2030813201 | 2063457736 | 2085946732 | 2100782021 |
| Spanish | 3 | 225360942 | 234880589 | 242979825 | 249825716 | 255425176 | 259665053 | 262506840 | 263913093 | 263920568 | 262649914 | 259748693 |
| Arabic | 4 | 199560003 | 221104972 | 242542805 | 264019933 | 285717708 | 307222944 | 327824786 | 347105695 | 364880316 | 381054179 | 398209563 |
| Malay | 5 | 31164177 | 33171611 | 35070236 | 36744569 | 38143162 | 39278935 | 40222737 | 41032165 | 41690545 | 42155191 | 42407851 |
| Russian | 6 | 143375006 | 142350640 | 140266843 | 137476901 | 134513277 | 131915337 | 129769608 | 127828308 | 125847236 | 123764841 | 121399781 |
| Bengali | 7 | 164827718 | 174038853 | 182177984 | 189043613 | 194521795 | 198600998 | 201279027 | 202562206 | 202545204 | 201301225 | 198406924 |
| Portuguese | 8 | 221508017 | 229026869 | 235340629 | 240353709 | 244028045 | 246370672 | 247430391 | 247296244 | 246042146 | 243701949 | 239599611 |
| French | 9 | 69804959 | 71745993 | 73641574 | 75573120 | 77484006 | 79276729 | 80924772 | 82494641 | 84072556 | 85698043 | 87745241 |
| Hausa | 10 | 415501622 | 468170800 | 524294294 | 584195666 | 647879029 | 714732720 | 783761326 | 854081578 | 924971069 | 995767407 | 1079747525 |
| Japanese | 11 | 126045211 | 124231042 | 121807156 | 118936515 | 115766736 | 112469341 | 109254758 | 106219925 | 103262690 | 100193980 | 96379129 |
| German | 12 | 89090207 | 89012844 | 88786378 | 88284805 | 87527607 | 86538908 | 85339890 | 83964218 | 82523245 | 81171585 | 79792478 |
| Persian | 13 | 80945718 | 84790653 | 87412320 | 89160309 | 90508508 | 91605731 | 92225509 | 92015992 | 90797505 | 88677567 | 85388791 |
| Swahili | 14 | 56877529 | 66049703 | 76249362 | 87618485 | 100150649 | 113706216 | 128121326 | 143282290 | 159083600 | 175425845 | 195558124 |
| Javaness | 15 | 263510146 | 277115590 | 289092102 | 299428362 | 308101650 | 314967329 | 320026086 | 323398215 | 325377255 | 326286943 | 326310161 |
| Korean | 16 | 50704971 | 51567916 | 52225667 | 52649343 | 52648814 | 52143644 | 51258205 | 50104988 | 48771226 | 47361790 | 45666602 |
| Turkish | 17 | 80417526 | 83325857 | 85972419 | 88850555 | 91435180 | 93524679 | 95117035 | 96187492 | 96754836 | 96827097 | 96328767 |
| Vietnamese | 18 | 95414640 | 99832970 | 103430834 | 106297805 | 108679883 | 110671471 | 112172245 | 113064212 | 113337381 | 113068857 | 112235584 |
| Italian | 19 | 59797978 | 59659281 | 59343986 | 58923116 | 58425413 | 57817603 | 57053322 | 56121000 | 55057288 | 53939675 | 52701460 |

Appendix2: Immigrant stock

## Immigrant stock by origin and destination

| | Chinese | English | Hindi | Spanish | Arabic | Malay | Russian | Bengali | Purtuguese | French | Hausa | Japanese | German | Persian | Swahili | Javaness | Korean | Turkish | Vietnamese | Italian |
|---|---|---|---|---|---|---|---|---|---|---|---|---|---|---|---|---|---|---|---|---|
| Chinese | 1381755664 | 4693209 | 7347 | 169605 | 723 | 10575 | 56270 | 166646 | 36579 | 112255 | 48736 | 741022 | 121799 | 0 | 0 | 73971 | 29042 | 2150 | 3141 | 203959 |
| English | 313027 | 91138890 | 65171 | 1385110 | 11835 | 427568 | 9694 | 321111 | 80665 | 306307 | 299364 | 336247 | 534267 | 0 | 36472 | 61785 | 4837 | 46051 | 17847 | 387199 |
| Hindi | 39621 | 5376245 | 1532920026 | 91446 | 1754 | 162747 | 6433 | 35250 | 11310 | 72931 | 68255 | 38411 | 143270 | 28117 | 10342 | 20664 | 537 | 1995 | 2643 | 225085 |
| Spanish | 407 | 13914782 | 0 | 210709734 | 507 | 0 | 1804 | 0 | 66387 | 349463 | 4173 | 4547 | 225874 | 0 | 0 | 0 | 0 | 1667 | 0 | 81597 |
| Arabic | 0 | 605554 | 0 | 8311 | 198686584 | 0 | 1332 | 0 | 4287 | 40213 | 13140 | 1754 | 56387 | 0 | 0 | 0 | 0 | 1108 | 0 | 141333 |
| Malay | 21775 | 344792 | 12317 | 681 | 2657 | 30546410 | 1557 | 208406 | 197 | 2575 | 1500 | 9099 | 8632 | 0 | 0 | 2315 | 197 | 311 | 133 | 623 |
| Russian | 3117 | 589566 | 1334 | 82329 | 7055 | 0 | 14139458 | 0 | 7473 | 64965 | 6028 | 8573 | 1104101 | 0 | 0 | 0 | 146 | 23567 | 0 | 82170 |
| Bengali | 4556 | 541734 | 3139311 | 10002 | 239 | 365600 | 471 | 160627571 | 1551 | 5069 | 4210 | 10073 | 9793 | 0 | 0 | 0 | 153 | 388 | 8254 | 98743 |
| Portuguese | 76940 | 730033 | 71 | 174021 | 138 | 0 | 560 | 15274 | 219998409 | 421601 | 88230 | 207193 | 283963 | 0 | 0 | 0 | 2807 | 525 | 0 | 108252 |
| French | 2228 | 575139 | 1130 | 223743 | 2991 | 0 | 3148 | 0 | 106665 | 68150157 | 103584 | 10135 | 325391 | 0 | 134988 | 0 | 0 | 19041 | 214 | 146405 |
| Hausa | 0 | 266731 | 0 | 1701 | 103 | 0 | 131 | 0 | 6500 | 2033 | 41208636 | 0 | 12665 | 0 | 516 | 0 | 0 | 170 | 0 | 2436 |
| Japanese | 21381 | 524464 | 708 | 10022 | 1086 | 16778 | 1025 | 14090 | 60429 | 21701 | 2993 | 125297180 | 41069 | 0 | 0 | 19345 | 212 | 2372 | 1926 | 8430 |
| German | 2740 | 1385296 | 2125 | 280130 | 8561 | 0 | 142706 | 0 | 67412 | 333893 | 107044 | 7434 | 85989271 | 0 | 1152 | 0 | 0 | 336904 | 0 | 425539 |
| Persian | 0 | 688461 | 3842 | 6453 | 141 | 0 | 2013 | 0 | 860 | 23134 | 1188 | 4533 | 147703 | 80033405 | 0 | 0 | 0 | 19299 | 0 | 14686 |
| Swahili | 0 | 119827 | 0 | 363 | 137 | 0 | 217 | 0 | 176 | 653 | 20001 | 0 | 921 | 0 | 56733340 | 0 | 0 | 262 | 0 | 1632 |
| Javanese | 177305 | 223775 | 1193 | 2237 | 2713 | 1091841 | 163 | 156500 | 1143 | 5021 | 2093 | 31068 | 23494 | 0 | 0 | 26177830 | 1428 | 211 | 8217 | 3214 |
| Korean | 195795 | 1451959 | 0 | 13113 | 281 | 3492 | 3723 | 0 | 10641 | 21016 | 2780 | 593449 | 33017 | 0 | 0 | 32677 | 48336536 | 605 | 1887 | 4000 |
| Turkish | 0 | 524224 | 103 | 4329 | 843 | 0 | 10244 | 0 | 744 | 301950 | 1896 | 0 | 1749222 | 2422 | 0 | 0 | 0 | 77800698 | 0 | 20851 |
| Vietnamese | 23941 | 903442 | 247 | 832 | 0 | 55295 | 8559 | 15216 | 127 | 56970 | 106 | 36791 | 60068 | 0 | 0 | 0 | 417 | 14 | 9250288 | 2327 |
| Italian | 652 | 1064083 | 0 | 115397 | 3939 | 0 | 1897 | 0 | 48315 | 373182 | 39925 | 3332 | 690523 | 0 | 685 | 0 | 0 | 3371 | 0 | 57452677 |

Appendix3: Soft power

## Soft Power Score

| language | No. | Soft Power |
|---|---|---|
| Mandarin Chinese | 0 | 50.5 |
| English | 1 | 293.79 |
| Hindustani | 2 | 30 |
| Spanish | 3 | 63.57 |
| Arabic | 4 | 30 |
| Malay | 5 | 30 |
| Russian | 6 | 49.6 |
| Bengali | 7 | 30 |
| Portuguese | 8 | 101.84 |
| French | 9 | 75.75 |
| Hausa | 10 | 30 |
| Japanese | 11 | 71.66 |
| German | 12 | 144.12 |
| Persian | 13 | 30 |
| Swahili | 14 | 30 |
| Javaness | 15 | 30 |
| Korean | 16 | 58.4 |
| Turkish | 17 | 45.35 |
| Vietnamese | 18 | 30 |
| Italian | 19 | 64.7 |

Appendix4: Original Distribution of L1 and L2

## Original Distribution

| each language speakers take up in total speaker | beta1(native) | beta2(second language) |
|---|---|---|
| Chinese(Madrin Wu Yue) | 0.241983852 | 0.094700687 |
| English | 0.085582468 | 0.299803729 |
| Hindustani(Telugu Tamil Marathi Punjabi) | 0.160322953 | 0.116781158 |
| Spanish | 0.100576701 | 0.044651619 |
| Arabic | 0.066897347 | 0.064769382 |
| Malay | 0.017762399 | 0.100098135 |
| Russian | 0.035294118 | 0.055446516 |
| Bengali | 0.055824683 | 0.009322866 |
| Portuguese | 0.050288351 | 0.005397448 |
| French | 0.017531719 | 0.075073602 |
| Hausa | 0.019607843 | 0.031894014 |
| Japanese | 0.029527105 | 0.000490677 |
| German | 0.017531719 | 0.025515211 |
| Persian | 0.01384083 | 0.029931305 |
| Swahili | 0.003690888 | 0.044651619 |
| Javanese | 0.019377163 | 0 |
| Korean | 0.017762399 | 0 |
| Turkish | 0.016378316 | 0 |
| Vietnamese | 0.015686275 | 0 |
| Italian | 0.014532872 | 0.001472031 |
| | alpha | 0.470126874 |

Appendix5: Language family classification

## Languages and their belonging language family

| Language zone | No | Sino-Tibetan | Indo-European | Afro-Asiatic | Austronesian | Niger–Congo | Dravidian | Turkic | Austroasiatic |
|---|---|---|---|---|---|---|---|---|---|
| Mandarin Chinese | 0 | 1 | 0 | 0 | 0 | 0 | 0 | 0 | 0 |
| English | 1 | 0 | 1 | 0 | 0 | 0 | 0 | 0 | 0 |
| Hindustani | 2 | 0 | 1 | 0 | 0 | 0 | 0 | 0 | 0 |
| Spanish | 3 | 0 | 1 | 0 | 0 | 0 | 0 | 0 | 0 |
| Arabic | 4 | 0 | 0 | 1 | 0 | 0 | 0 | 0 | 0 |
| Malay | 5 | 0 | 0 | 0 | 1 | 0 | 0 | 0 | 0 |
| Russian | 6 | 0 | 1 | 0 | 0 | 0 | 0 | 0 | 0 |
| Bengali | 7 | 0 | 1 | 0 | 0 | 0 | 0 | 0 | 0 |
| Portuguese | 8 | 0 | 1 | 0 | 0 | 0 | 0 | 0 | 0 |
| French | 9 | 0 | 1 | 0 | 0 | 0 | 0 | 0 | 0 |
| Hausa | 10 | 0 | 0 | 1 | 0 | 0 | 0 | 0 | 0 |
| Japanese | 11 | 0 | 0 | 0 | 0 | 0 | 0 | 0 | 0 |
| German | 12 | 0 | 1 | 0 | 0 | 0 | 0 | 0 | 0 |
| Persian | 13 | 0 | 1 | 0 | 0 | 0 | 0 | 0 | 0 |
| Swahili | 14 | 0 | 0 | 0 | 0 | 1 | 0 | 0 | 0 |
| Javaness | 15 | 0 | 0 | 0 | 0 | 0 | 1 | 0 | 0 |
| Korean | 16 | 0 | 0 | 0 | 0 | 0 | 0 | 0 | 0 |
| Turkish | 17 | 0 | 0 | 0 | 0 | 0 | 0 | 1 | 0 |
| Vietnamese | 18 | 0 | 0 | 0 | 0 | 0 | 0 | 0 | 1 |
| Italian | 19 | 0 | 1 | 0 | 0 | 0 | 0 | 0 | 0 |

Appendix6: Export between language zones

| Language zone | Chinese | English | Hindustani | Spanish | Arabic | Malay | Russian | Bengali | Portuguese | French | Hausa | Japanese | German | Persian | Swahili | Javanese | Korean | Turkish | Vietnamese | Italian |
|---|---|---|---|---|---|---|---|---|---|---|---|---|---|---|---|---|---|---|---|---|
| Chinese | 0.231721 | 0.319618 | 0.0104789 | 0.06065 | 0 | 0.004337 | 0.011158 | 0.003079 | 0.00950126 | 0.040312 | 0 | 0.035882 | 0.143642 | 0.004861 | 0 | 0.007564 | 0.027126 | 0.01863 | 0.018273515 | 0.053166 |
| English | 0.185931 | 0.348444 | 0.0233375 | 0.188808 | 0 | 0.001215 | 0 | 0 | 0.02754325 | 0.042327 | 0 | 0.065187 | 0.067326 | 0 | 0 | 0.00147 | 0.043243 | 0.003862 | 0.001022741 | 0.000284 |
| Hindustani | 0.304918 | 0.603279 | 0 | 0 | 0 | 0 | 0 | 0 | 0 | 0 | 0 | 0 | 0.091803 | 0 | 0 | 0 | 0 | 0 | 0 | 0 |
| Spanish | 0.031658 | 0.52214 | 0.0103006 | 0.034776 | 0 | 0 | 0 | 0 | 0.06509081 | 0.125059 | 0 | 0.007287 | 0.109222 | 0 | 0 | 0 | 0.003083 | 0.023004 | 0 | 0.068378 |
| Arabic | 0.231721 | 0.319618 | 0.0104789 | 0.06065 | 0 | 0.004337 | 0.011158 | 0.003079 | 0.00950126 | 0.040312 | 0 | 0.035882 | 0.143642 | 0.004861 | 0 | 0.007564 | 0.027126 | 0.01863 | 0.018273515 | 0.053166 |
| Malay | 0.366726 | 0.266547 | 0 | 0 | 0 | 0 | 0 | 0 | 0 | 0 | 0 | 0.144902 | 0.051878 | 0 | 0 | 0.06619 | 0.051878 | 0 | 0.051873354 | 0 |
| Russian | 0.26972 | 0.142494 | 0 | 0 | 0 | 0 | 0 | 0 | 0 | 0 | 0 | 0.07888 | 0.180662 | 0 | 0 | 0 | 0.096692 | 0.129771 | 0 | 0.101781 |
| Bengali | 0.231721 | 0.319618 | 0.0104789 | 0.06065 | 0 | 0.004337 | 0.011158 | 0.003079 | 0.00950126 | 0.040312 | 0 | 0.035882 | 0.143642 | 0.004861 | 0 | 0.007564 | 0.027126 | 0.01863 | 0.018273515 | 0.053166 |
| Portuguese | 0.387126 | 0.202705 | 0.033873 | 0.095061 | 0 | 0.029254 | 0.020016 | 0.036952 | 0.00191201 | 0.018532 | 0 | 0.032333 | 0.052111 | 0 | 0 | 0 | 0.020016 | 0 | 0.03849202 | 0.031616 |
| French | 0.081416 | 0.240708 | 0 | 0.134513 | 0 | 0 | 0 | 0 | 0.01946903 | 0 | 0 | 0.023009 | 0.338053 | 0 | 0 | 0 | 0 | 0.024779 | 0 | 0.138053 |
| Hausa | 0.231721 | 0.319618 | 0.0104789 | 0.06065 | 0 | 0.004337 | 0.011158 | 0.003079 | 0.00950126 | 0.040312 | 0 | 0.035882 | 0.143642 | 0.004861 | 0 | 0.007564 | 0.027126 | 0.01863 | 0.018273515 | 0.053166 |
| Japanese | 0.445469 | 0.376344 | 0 | 0.026114 | 0 | 0 | 0 | 0 | 0 | 0 | 0 | 0 | 0.041475 | 0 | 0 | 0 | 0.110599 | 0 | 0 | 0 |
| German | 0.198788 | 0.415591 | 0.0181746 | 0.005553 | 0 | 0 | 0.052575 | 0 | 0 | 0.013631 | 0 | 0.006563 | 0.146801 | 0 | 0 | 0 | 0 | 0 | 0 | 0.142323 |
| Persian | 0.231721 | 0.319618 | 0.0104789 | 0.06065 | 0 | 0.004337 | 0.011158 | 0.003079 | 0.00950126 | 0.040312 | 0 | 0.035882 | 0.143642 | 0.004861 | 0 | 0.007564 | 0.027126 | 0.01863 | 0.018273515 | 0.053166 |
| Swahili | 0.231721 | 0.319618 | 0.0104789 | 0.06065 | 0 | 0.004337 | 0.011158 | 0.003079 | 0.00950126 | 0.040312 | 0 | 0.035882 | 0.143642 | 0.004861 | 0 | 0.007564 | 0.027126 | 0.01863 | 0.018273515 | 0.053166 |
| Javanese | 0.231721 | 0.319618 | 0.0104789 | 0.06065 | 0 | 0.004337 | 0.011158 | 0.003079 | 0.00950126 | 0.040312 | 0 | 0.035882 | 0.143642 | 0.004861 | 0 | 0.007564 | 0.027126 | 0.01863 | 0.018273515 | 0.053166 |
| Korean | 0.372881 | 0.289676 | 0.0400616 | 0.029276 | 0 | 0.021572 | 0 | 0 | 0 | 0 | 0 | 0.072419 | 0.023112 | 0 | 0 | 0.023112 | 0 | 0 | 0.12788906 | 0 |
| Turkish | 0 | 0.297222 | 0 | 0.108333 | 0 | 0 | 0 | 0 | 0 | 0.119444 | 0 | 0 | 0.261111 | 0.058333 | 0 | 0 | 0 | 0 | 0 | 0.155556 |
| Vietnamese | 0.231721 | 0.319618 | 0.0104789 | 0.06065 | 0 | 0.004337 | 0.011158 | 0.003079 | 0.00950126 | 0.040312 | 0 | 0.035882 | 0.143642 | 0.004861 | 0 | 0.007564 | 0.027126 | 0.01863 | 0.018273515 | 0.053166 |
| Italian | 0.136015 | 0.130268 | 0 | 0.105364 | 0 | 0 | 0.061303 | 0 | 0 | 0.164751 | 0 | 0 | 0.360153 | 0 | 0 | 0 | 0 | 0.042146 | 0 | 0 |

Appendix7: Foreign Direct Investment

| | FDI |
|---|---|
| Language zone | Foreign direct investment net outflows (% of GDP) |
| Chinese | 2.5053676 |
| English | 1.73414805 |
| Hindustani | 0.200545931 |
| Spanish | 2.35731272 |
| Arabic | 0 |
| Malay | 3.393477038 |
| Russian | 1.739009794 |
| Bengali | 0.018266995 |
| Portuguese | 0.916805772 |
| French | 2.581593114 |
| Hausa | 0.722097745 |
| Japanese | 3.434178695 |
| German | 2.798835114 |
| Persian | 0 |
| Swahili | 0.024849195 |
| Javanese | -1.265879213 |
| Korean | 1.932633143 |
| Turkish | 0.364357686 |
| Vietnamese | 0 |
| Italian | 0.819745164 |